\documentclass{article}

\PassOptionsToPackage{square, numbers}{natbib}
\usepackage[final]{nips_2017}

\usepackage[utf8]{inputenc} 
\usepackage[T1]{fontenc}    
\usepackage{hyperref}       
\usepackage{url}            
\usepackage{booktabs}       
\usepackage{amsfonts, amsmath}       
\usepackage{nicefrac}       
\usepackage{microtype}      
\usepackage[pdftex]{graphicx}

\title{Ensemble Multi-task Gaussian Process Regression with Multiple Latent Processes}

\author{
  Weitong Ruan, Eric L. Miller \\
  Dept. of Electrical and Computer Engineering\\
  Tufts University \\
  Medford, MA, 02155 \\
  \texttt{weitong.ruan@tufts.edu, elmiller@ece.tufts.edu}
}

\begin{document}

\maketitle

\begin{abstract}
	Multi-task/Multi-output learning seeks to exploit correlation among tasks to enhance performance over learning or solving each task independently. In this paper, we investigate this problem in the context of Gaussian Processes (GPs) and propose a new model which learns a mixture of latent processes by decomposing the covariance matrix into a sum of structured hidden components each of which is controlled by a latent GP over input features and a "weight" over tasks. From this sum structure, we propose a parallelizable parameter learning algorithm with a predetermined initialization for the "weights". We also notice that an ensemble parameter learning approach using mini-batches of training data not only reduces the computation complexity of learning but also improves the regression performance. We evaluate our model on two datasets, the smaller Swiss Jura dataset and another relatively larger ATMS dataset from NOAA. Substantial improvements are observed compared with established alternatives.
\end{abstract}

\section{Introduction}
Compared with the traditional supervised learning problem, where the output is a scalar, multi-task/multi-output problems are characterized by vector-valued outputs. The application of multi-task learning ranges from sensor network estimation to resource estimation in environmental applications and phenotype prediction in genetics \cite{rakitsch2013all}. Within the context of multi-task GP, one direction is to reduce these problems to one of kernel design and learning \cite{bonilla2007multi, alvarez2012kernels}, where a kernel function is the product of a task covariance function and a sample covariance function. The task covariance function can be either parameterized as in \cite{bonilla2007kernel} or "free-form" as in \cite{bonilla2007multi} and the resulting covariance matrix becomes a Kronecker product of a "task-similarity" matrix and a matrix capturing the covariance structure of the samples along with the addition of a diagonal matrix to account for noise. In terms of viewing each task/output as a mixture of latent processes, this model corresponds to a scenario where only one latent process exists and each task simply shares this latent process with various weights plus additive white noise. 

In this paper, we propose a model that is optimized from the so-called "no transfer" model, i.e. learning tasks separately with independent GPs. From a generative modeling perspective, in this "no transfer" model, each task is generated solely from one latent processes with additive white noise. In our model, we inherit the assumption that for each task there exists a latent process, but we improve the previous model by allowing each task to be generated from a mixture of all latent processes where the "weights" are learned from the observed data. In terms of the covariance matrix, the limited single Kronecker product structure is replaced with a sum of Kronecker products of "weights" and covariance matrices capturing correlation among samples obtained from each latent process, which greatly expands the expressiveness of the model. 

To solve the problem of increased number of parameters and computation complexity, we developed a two-step parameter learning algorithm, both of which can be executed in parallel to reduce computation time. The first step is to learn multiple latent processes, one for each task. Then we learn the "weights" using data from all tasks. Parallelization for the first step can be easily implemented by learning the parameters for each latent process simultaneously. Then for the second step, we divide the training dataset from all tasks into mini-batches, resulting in an ensemble of regressors, one over each mini-batch. The final regression result becomes an average of results from all regressors. We notice that, although the original purpose of partition training data into mini-batches is to reduce computational complexity, in some cases we observe that it can, unexpectedly, boost performance.

In addition to the fact that our parameter learning algorithm can work in parallel, another advantage is the ability to incorporate recently developed GP learning algorithms. More specifically, because the initial stage of our approach requires the learning of independent GP processes, future efforts would allow for the exploitation of any of several new approaches developed in recent years for solving GP learning problems notable for their reduced computational complexity or computation time. Hensman et al introduced a stochastic variational inference (SVI) based sparse GP algorithm in \cite{hensman2013gaussian}. Gal et al provided a re-parameterization of variational inference for sparse GP and GPLVM, which allows for an efficient distributed algorithm \cite{gal2014distributed} and Deisenroth et al proposed a Product-of-GP-experts models (PoEs) based distributed GP algorithm that can scale GP to an arbitrarily large dataset \cite{deisenroth2015distributed}. To the best of our knowledge, the approach we propose here represents the first method capable of employing these sophisticated single task GP learning ideas for multi-task problems. 

The paper is structured as follows: in Section 2, we showed that how our model is optimized from the "no-transfer" model and the ensemble approach is detailed in Section 3. Related work is described in Section 4 and some experimental results are presented in Section 5.

\section{The Model}
\label{Model}

Let the matrix\footnote{Throughout this paper, we use bold symbols for both matrices (capitalized) and vectors (lower case).} $\boldsymbol{Y} = \begin{bmatrix}
\boldsymbol{y}_1 & \boldsymbol{y}_2 & \dotsb & \boldsymbol{y}_D
\end{bmatrix} \in \mathbb{R}^{N \times D}$ be a set of responses with $N$ samples and $D$ tasks, where each column $\boldsymbol{y}_d$ represents the $d$-th task with length $N$. We also use $\boldsymbol{y} = vec \left( \boldsymbol{Y} \right) = \begin{bmatrix}
\boldsymbol{y}_1^T & \dotsb & \boldsymbol{y}_D^T
\end{bmatrix}^T \in \mathbb{R}^{ND \times 1}$  and denote by $\boldsymbol{X}$ a set of $N$ inputs $\boldsymbol{x}_1, \boldsymbol{x}_2, \cdots, \boldsymbol{x}_N$. The multi-task regression problem is with a new set of $M$ inputs $\boldsymbol{X}^*$, predict their vector-value responses $\boldsymbol{y}^*$.

In the "no transfer" model, each task is generated solely from one latent process. For simplicity, we assume that each task is normalized with zero mean and unit variance:
\begin{equation}
\boldsymbol{y}_d \sim \mathcal{N} \left( \boldsymbol{0}, \boldsymbol{K}_d + \sigma_d^2 \boldsymbol{I} \right) \qquad \qquad \forall d = 1, 2, \cdots, D.
\end{equation}
As detailed in the supplemental material, the resulting covariance matrix of $\boldsymbol{y}$ can be written as:
\begin{equation}
\label{BasicCov}
\boldsymbol{K_y} = \sum_{d = 1}^{D} \boldsymbol{B}_d \otimes \boldsymbol{K}_d + \boldsymbol{D} \otimes \boldsymbol{I} \qquad \qquad  \boldsymbol{D} = diag \left( \sigma_1^2, \sigma_2^2, \cdots, \sigma_D^2 \right)
\end{equation}
where $\otimes$ stands for the Kronecker product, $\boldsymbol{B}_d$, a $D \times D$ matrix of zeros except for a single 1 as $d$-th element along the diagonal, $\boldsymbol{K}_d$ corresponds to the covariance matrix for the data associated with the $d$-th task, and $\boldsymbol{D}$ is a $D \times D$ diagonal matrix where the $d$-th element on the diagonal corresponds to the noise variance $\sigma_d^2$ of the $d$-th task. Note, the matrix $\boldsymbol{B}_d$ is referred to by different names in various papers, in our model, we call it the "weights", which originates from the generative perspective that data are generated from a mixture of latent processes.

Inspired by the covariance structure in equation \eqref{BasicCov}, in our model, we allow $\boldsymbol{B}_d$ to be a dense matrix. Also with the assumption that $\boldsymbol{B}_d$ is positive semi-definite (PSD), certainly true in \eqref{BasicCov}, here we write $\boldsymbol{B}_d$ more generally as $\boldsymbol{B}_d = \boldsymbol{W}_d\boldsymbol{W}_d^T$, a matrix product or as a rank-$k$ approximation $\boldsymbol{B}_d \approx \sum_{j=1}^{k} \boldsymbol{w}_d^j \left( \boldsymbol{w}_d^j \right)^T$. In practice, a rank-$k$ approximation is often used to avoid overfitting \cite{rakitsch2013all}. Thus, in our model, the covariance matrix can be written as or approximated as:
\begin{equation}
\label{KeyCov}
\boldsymbol{K_y} = \sum_{d = 1}^{D} \boldsymbol{W}_d \boldsymbol{W}_d^T \otimes \boldsymbol{K}_d + \boldsymbol{D} \otimes \boldsymbol{I} \approx \sum_{d = 1}^{D} \left( \sum_{j=1}^{k} \boldsymbol{w}_d^j \left( \boldsymbol{w}_d^j \right)^T \right) \otimes \boldsymbol{K}_d + \boldsymbol{D} \otimes \boldsymbol{I}
\end{equation}
Compared with the "no-transfer" model, the advantage of our model is each covariance block is a mixture of matrices $\{\boldsymbol{K}_d\}_{d=1}^D$, in other words, each task is generated from a mixture of multiple latent processes and the "weights" \{$\boldsymbol{W}_d\}_{d=1}^D$ or $\{\{\boldsymbol{w}_d^j\}_{j=1}^k\}_{d=1}^D$ are learned from observed data. Note in this paper, we simply use $k=1$, which is a rank-1 approximation. We emphasize here the choice of $D$ in our model is motivated entirely by the structure of \eqref{BasicCov}, which poses an interesting future work of exploring generalization or automatic methods for selecting the optimal $D$. 

\subsection{Inference}
Once the covariance structure is set as in equation \eqref{KeyCov}, inference in our model follows the standard GP framework \cite{williams2006gaussian}. The predictive distribution for unseen set of inputs $\boldsymbol{X}^*$:
\begin{equation}
p \left( \boldsymbol{y}^* \lvert \boldsymbol{y}, \{ \{\boldsymbol{w}_d^j\}_{j=1}^k, \boldsymbol{\alpha}_d, \sigma_d \}_{d=1}^D\right) = \mathcal{N} \left( \boldsymbol{K_{y*y}} \left( \boldsymbol{K_y} \right)^{-1} \boldsymbol{y} , \boldsymbol{K_{y*}}  - \boldsymbol{K_{y*y}} \left( \boldsymbol{K_y} \right)^{-1} \boldsymbol{K_{yy*}}  \right)
\end{equation}
where $\boldsymbol{\alpha}_d$ is a vector of parameters controlling the $d$-th latent process, $\boldsymbol{K_y}, \boldsymbol{K_{y*}}$ stand for the autocovariance matrices of $\boldsymbol{y}$ and $\boldsymbol{y}^*$ and $\boldsymbol{K_{y*y}}, \boldsymbol{K_{yy*}}$ are cross-covariance matrices between training and test instances.

\subsection{Parameter learning}
\label{parameterLearning}
The parameters in our model that need to be learned from training data are $\{ \{\boldsymbol{w}_d^j\}_{j=1}^k, \boldsymbol{\alpha}_d, \sigma_d \}_{d=1}^D$. Compared with existing approaches \cite{rakitsch2013all, bonilla2007multi, alvarez2011computationally, wilson2012gaussian}, where a unified learning algorithm is designed and matrix approximations (e.g, incomplete-Cholesky decomposition) are often used to reduce computation complexity, in our model, parameter learning can be implemented in a straight-forward two step process, without the need for a complex derivation of the lower bound of the marginal likelihood typically required by EM or variational learning.

We first learn $\boldsymbol{\alpha}_d$ and $\sigma_d$, parameters that control the $d$-th latent process, from the $d$-th task $\boldsymbol{y}_d$, independently from other tasks, by maximizing the log marginal likelihood of data from each task $\log p \left( \boldsymbol{y}_d \lvert \boldsymbol{X}, \boldsymbol{\alpha}_d, \sigma_d \right)$ 
\begin{equation}
\hat{\boldsymbol{\alpha}}_d, \hat{\sigma}_d = \arg \max_{\boldsymbol{\alpha}_d, \sigma_d} -\frac{1}{2} \boldsymbol{y}_d^T \boldsymbol{K}_d^{-1} \boldsymbol{y}_d - \frac{1}{2} \log \lvert \boldsymbol{K}_d \rvert - \frac{N}{2} \log \left( 2 \pi \right).
\end{equation}
In standard GP, this optimization problem is solved via an iterative gradient-based ascent algorithm, where every element of the gradient can be easily calculated as
\begin{equation}
\label{step1Grad}
\frac{\partial}{\partial \theta} \log p \left( \boldsymbol{y}_i \lvert \boldsymbol{X}, \boldsymbol{\alpha}_d, \sigma_d \right) = \frac{1}{2} \text{tr} \left( \left( \boldsymbol{\gamma} \boldsymbol{\gamma}^T - \boldsymbol{K}_d^{-1} \right) \frac{\partial \boldsymbol{K}_d}{\partial \theta} \right)
\end{equation}
where $\boldsymbol{\gamma} = \boldsymbol{K}_d^{-1} \boldsymbol{y}_d$ and $\theta$ can be any variable in $\boldsymbol{\alpha}_d$ or $\sigma_d$. The same algorithm is repeated $D$ times to learn the entire set $ \{ \boldsymbol{\alpha}_d, \sigma_d \}_{d=1}^D$. Note this step is the same as parameter learning in the "no transfer" model, finding $D$ independent latent processes independently from each task. 

Secondly, we learn the set of weight vectors $\{ \{\boldsymbol{w}_d^j\}_{j=1}^k \}_{d=1}^D$ jointly from the whole training data $\boldsymbol{y}$ by maximizing the log marginal likelihood of the entire training dataset $\log p \left( \boldsymbol{y} \lvert \boldsymbol{X}, \{ \{\boldsymbol{w}_d^j\}_{j=1}^k, \hat{\boldsymbol{\alpha}}_d, \hat{\sigma}_d \}_{d=1}^D\right)$, where in place of the $ \{ \boldsymbol{\alpha}_d, \sigma_d \}_{d=1}^D$, we use the estimates obtained in the last step to build the matrix $\boldsymbol{K_y}$;
\begin{equation}
 \{ \{\hat{\boldsymbol{w}}_d^j\}_{j=1}^k \}_{d=1}^D = \arg \max_{\{ \{\boldsymbol{w}_d^j\}_{j=1}^k \}_{d=1}^D} -\frac{1}{2} \boldsymbol{y}^T \boldsymbol{K_y}^{-1} \boldsymbol{y} - \frac{1}{2} \log \lvert \boldsymbol{K_y} \rvert - \frac{ND}{2} \log \left( 2 \pi \right)
\end{equation}
where $\boldsymbol{K_y} $ depends on $\boldsymbol{w}_d^j$ as in equation \eqref{KeyCov}.
This problem is again solved using an iterative gradient based ascent algorithm and the gradient has a similar form as in previous steps with minor changes,
\begin{equation}
\frac{\partial}{\partial \theta} \log p \left( \boldsymbol{y} \lvert \boldsymbol{X}, \{ \{\boldsymbol{w}_d^j\}_{j=1}^k, \hat{\boldsymbol{\alpha}}_d, \hat{\sigma}_d \}_{d=1}^D\right) = \frac{1}{2} \text{tr} \left( \left( \boldsymbol{\gamma} \boldsymbol{\gamma}^T - \boldsymbol{K_y}^{-1} \right) \frac{\partial \boldsymbol{K_y}}{\partial \theta} \right),
\end{equation} 
\begin{equation}
\frac{\partial \boldsymbol{K_y}}{\partial \theta} = \left( \boldsymbol{w}_d^j \boldsymbol{\delta}_l^T + \boldsymbol{\delta}_l \boldsymbol{w}_d^{jT} \right) \otimes \boldsymbol{K}_d
\end{equation}
where $\boldsymbol{\gamma} = \boldsymbol{K_y}^{-1} \boldsymbol{y}$, $\boldsymbol{\delta}_l$ is a Kronecker delta vector and $\boldsymbol{\theta}$ is the $l$-th element of weight $\boldsymbol{w}_d^j$. Details regarding the calculation of gradients are presented in the supplemental material.

As with previous algorithms, the main problem of computational complexity for multi-task GP is the inversion of the covariance matrix $\boldsymbol{K_y}$ in equation \eqref{KeyCov} \cite{alvarez2012kernels}. Notice that this covariance matrix $\boldsymbol{K_y}$ is of size $ND \times ND$, hence its inverse needs at least $O \left( D^3 N^3 \right)$ in time. Compared with other methods, reviewed in \cite{alvarez2012kernels}, developed to reduce complexity, the idea of our approach is to partition data in a clever and reasonable manner and solve several smaller matrix inversions in parallel instead of a large matrix inversion. 

In the first step of our learning algorithm, each set of parameters $ \{ \boldsymbol{\alpha}_d, \sigma_d \}$ are only learned from each task $\boldsymbol{y}_d$, reducing the size of covariance matrix needed from  $ND \times ND$ to $N \times N$ and the time complexity to $O \left( N^3 \right)$. Although this estimation needs to run $D$ times, since each set of parameters are learned independently for each task, this step can be implemented in parallel. Note these calculations can be further accelerated by applying sparse GP approximation techniques designed for standard one output GP, for example, DTC \cite{csato2001sparse, seeger2003fast, quinonero2005unifying}, PITC \cite{quinonero2005unifying}, FITC \cite{snelson2006sparse}, where selective inducing samples of size $M << N$ are used for parameter learning and making predictions with complexity $O \left( M^2N \right)$, as well as more recent approaches developed for large-scale GP. In Section \ref{SecEnsemble}, we detail another mechanism that partition the training data into a set of mini-batches, on which we train an ensemble of regressors; each with shared $\{\hat{\boldsymbol{\alpha}}_d, \hat{\sigma}_d \}_{d=1}^D$, but different $\{ \boldsymbol{w}_d^j \}_{d=1}^D$. Due to this data partition, although we consider all tasks simultaneously, the reduction in $N$ brings down the computation complexity.

\subsection{Another interpretation of parameter learning algorithm}
\label{AnotherInterpolation}

The above parameter learning algorithm can also be explained from a more general perspective where we always wish to maximize the log marginal likelihood of the entire dataset $\log p \left( \boldsymbol{y} \lvert \{ \{\boldsymbol{w}_d^j\}_{j=1}^k, \boldsymbol{\alpha}_d, \sigma_d \}_{d=1}^D\right)$. For example, in the case when $k=1$, the objective function simplifies as $\log p \left( \boldsymbol{y} \lvert \{ \boldsymbol{w}_d, \boldsymbol{\alpha}_d, \sigma_d \}_{d=1}^D\right)$ and the parameters are learned in a cyclic coordinate ascent manner. The algorithm starts from $\boldsymbol{w}_d = \boldsymbol{\delta}_d$, where $\boldsymbol{\delta}_d$ is a Kronecker delta vector, optimizes over $\{\boldsymbol{\alpha}_d, \sigma_d \}_{d=1}^D$ and then use the estimated $\{\hat{\boldsymbol{\alpha}}_d, \hat{\sigma}_d \}_{d=1}^D$ to maximize the objective function with respect to $\{ \boldsymbol{w}_d \}_{d=1}^D$. The algorithm described in Section \ref{parameterLearning} stops after the first iteration, while from this perspective, we notice that the parameter can be further optimized in this manner, however, after the first iteration, due to the dense property of the optimized "weights", each task becomes correlated and the parallel learning for $\{\boldsymbol{\alpha}_d, \sigma_d \}_{d=1}^D$ needs to be modified, which falls out of the context of this paper, so we leave it to future work.

Note that in the case when $k=1$, $\boldsymbol{\delta}_d$ becomes a predetermined initialization for the weight $\boldsymbol{\delta}_d$. To demonstrate that $\{ \boldsymbol{\delta}_d \}_{d=1}^D$ provides a valid initialization, consider the case where data is generated from only one latent process. The covariance matrix in this case has only one Kronecker product as $\boldsymbol{K_y} = \boldsymbol{B} \otimes \boldsymbol{K} + \boldsymbol{D} \otimes \boldsymbol{I}$ and this corresponds to the widely used ICM model \cite{bonilla2007multi, alvarez2012kernels}. In our model, one latent process is equivalent to all $\boldsymbol{K}_d$ matrices being the same. Suppose $\boldsymbol{K}_d = \boldsymbol{K}$ for all $k = 1, \cdots, D$. If we use the rank-1 approximation form in equation \eqref{KeyCov}, we have
\begin{equation}
\label{SVD}
\sum_{d = 1}^{D} \boldsymbol{w}_d\left( \boldsymbol{w}_d \right)^T  \otimes \boldsymbol{K}_d =  \sum_{d = 1}^{D}  \boldsymbol{w}_d \left( \boldsymbol{w}_d \right)^T  \otimes \boldsymbol{K}  = \boldsymbol{U} \boldsymbol{\Sigma} \boldsymbol{U}^T \otimes \boldsymbol{K}
\end{equation}
where $\boldsymbol{U} = \begin{bmatrix}
\boldsymbol{u}_1 & \dotsb & \boldsymbol{u}_D
\end{bmatrix}$ is a unitary matrix with the $i$-th column being a unit vector $\boldsymbol{u}_d$, $\boldsymbol{\Sigma}$ is a diagonal matrix with the $i$-th element on the diagonal being $\lVert \boldsymbol{w}_d \rVert ^2$. Due to the fact that matrix $\boldsymbol{B}$ is assumed to be positive semidefinite, $\boldsymbol{U} \boldsymbol{\Sigma} \boldsymbol{U}^T$ can be treated as a singular value decomposition of $\boldsymbol{B}$, which requires $\boldsymbol{U}$ to be an orthonormal matrix. It's obvious that $\{ \boldsymbol{\delta}_d \}_{d=1}^D$ forms an orthonormal basis, thus it is a valid starting point. From the above reasoning, we also notice that the ICM model with a full rank $\boldsymbol{B}$ corresponds to a special case of our model where only one latent process exists.

\section{The Ensemble Approach}
\label{SecEnsemble}
In Section \ref{parameterLearning}, we demonstrated that in our two-step parameter learning algorithm, the first step can be easily separated into $D$ parallelizable standard GP parameter learning subproblems. To expedite the second step, we partition the training data into $L$ mini-batches $P_1, \cdots P_L$ of size $N_0$, where $L = \lfloor \frac{N}{N_0} \rfloor$,  and train an ensemble of $L$ different sets of weight vectors $\{ \boldsymbol{w}_d \}_{d=1}^D$. The quantity $N_0$ can either be a constant, or grows as a function of $D$. In this paper, we always choose $N_0 = D^2$. We also notice that the choice of $N_0$ influences regression performance, but we leave this discussion to Section \ref{ExpJura}, where the effects of ensemble will be explored using real data.

Suppose the training instances are indexed from $1$ to $N$, then instances indexed by $\{ k, L+k, 2*L+k, \cdots,  (N_0-1)*L+k\}$ are partitioned into mini-batch $P_k$. In terms of computational cost, due to the data partition, we again divide the second step of our learning algorithm into $L$ parallelable subproblems. In each subproblem, instead of inverting a large matrix of size $ND \times ND$, we only need to deal with a smaller matrix of size $N_0D \times N_0D$. 


With training data being partitioned into $L$ mini-batches, we are able to learn an ensemble of $L$ regressors, and then we average the result from each regressor to form the final regression result. Note other methods, for example, the PoEs model in \cite{deisenroth2015distributed}, can also be developed to possibly better incorporate results from the ensemble, we choose the simple average just to demonstrate the effectiveness of ensemble approaches and this is different from averaging $L$ set of weight vectors $\{ \boldsymbol{w}_d \}_{d=1}^D$ directly since the regression result is not linear w.r.t the "weights".

\section{Related Work}
The use of Gaussian Processes for multi-task learning problems was mainly developed in the geostatistics and machine learning communities. Most of the previous work fall in the category of using one or more Kronecker products to model the covariance matrix \cite{rakitsch2013all, bonilla2007kernel, goovaerts1997geostatistics, journel1978mining, teh2005semiparametric}. In geostatistics, covariance matrices with more than one Kronecker products are referred to as "linear model of coregionalization" (LMC) \cite{goovaerts1997geostatistics, journel1978mining} and the simplified version "intrinsic coregionalization model" (ICM) features only one Kronecker product. Our model is a special case of LMC where $D$ Kronecker products is used. As noted in \cite{journel1978mining}, the ICM model is much more restrictive than the LMC model. This can also be seen from equation \eqref{SVD} in Section \ref{AnotherInterpolation}, even with a full-rank $\boldsymbol{B}$, ICM is just a special case of our model ($k=1$) when all latent processes are the same and the weight vectors are left-singular vectors scaled by the square root of corresponding singular values.

In machine learning, one of the early ideas that is proposed to solve multi-task GP used the limited ICM model with $\boldsymbol{B}$ being an identity matrix and the correlation is imposed implicitly by assuming the same set of parameters for all tasks \cite{lawrence2004learning, yu2005learning}. Bonilla et al \cite{bonilla2007multi} used the full ICM model, where $\boldsymbol{B}$ needs not to be an identity matrix, with an EM-based algorithm for parameter learning. Rakitsch et al extended the ICM model by allowing noise to be correlated and proposed an efficient parameter estimation and inference algorithm by taking advantage of the two Kronecker products structure. The semiparameric latent factor models (SLFM) involved using the LMC model with rank-$1$ approximation to each $\boldsymbol{B}_i$. The optimal number of latent processes are obtained by naively trying each possible number and choose the one that gives the largest marginal log likelihood.

Mauricio and Neil went beyond the separable kernels by proposing to use a convolution process with a smoothing function to provide a non-instantaneous mixing of latent processes \cite{alvarez2011computationally}. Another direction to extend the separable kernel follows the idea to allow "task-similarity" matrix to vary as functions of input features, e.g. the Gaussian Process Regression Networks in \cite{wilson2012gaussian} which employs a Bayesian neural network to provide an adaptive mixing of latent processes and noise correlations. 

\section{Experiments}
\label{Exp}
We evaluate the performance of our model using two real datasets. The first dataset contains the concentration of several metal pollutants in a region of the Swiss Jura and the job is to predict the concentration of one or some metal pollutants at test locations given training data from observed training locations. This dataset appears as a standard dataset to evaluate multi-task regression performance in e.g. \cite{alvarez2011computationally, wilson2012gaussian}.  The second dataset is collected by Advanced Technology Microwave Sounder (ATMS) from NOAA Comprehensive Large Array-data Stewardship System (CLASS) database. Because of the geometry of satellite data acquisition, these data are not collected on a regular grid, however, in meteorological applications, gridded data are often preferred in order to generate subsequent scientific products, such as vertical temperature and water vapor profiles. This task is further complicated by the fact that for each locations (inputs), ATMS measures several frequency channels in bands from 23 GHz to 183 GHz, hence a multi-task regression is needed to infer a vector value output on a given location on a grid (test input).

Throughout the experiments, the GP module from \textit{scikit-learn} \cite{scikit-learn} is used to implement GP learning for each task and a variant of \textit{L-BFGS} \cite{zhu1997algorithm} is used for all optimizations involved. The maximum number of iterations is set to 120. The covariance function that we use is the squared exponential (SE) with a global length-scale. Unless otherwise noted, data from each task are first normalized to zero mean and unit variance before the learning algorithm is applied.

\subsection{Jura Dataset}
\label{ExpJura}
This dataset contains the concentration of seven metals at 359 sample sites with their corresponding spatial coordinates\footnote{This data is available at https://sites.google.com/site/goovaertspierre/pierregoovaertswebsite/download}. It is further partitioned into a training set of 259 samples and a test set of 100 samples. In this experiment, we are interested in predicting the concentration of Cadmium at testing sites given its concentration at training sites. In addition, we also have access to the concentration of zinc and nickel at both training and testing sites with the hope that correlations among those three metals can be learned from training set and use this correlation with zinc and nickel concentration on the test sites to improve the prediction performance. From the perspective of our model, the spatial correlation for the concentration of each metal can be modeled using a separate GP, then their correlation is learned from the training data through the optimized "weights". Although we aim to predict only the concentration of Cadmium, resulting in a scalar output, the correlation is modeled using multi-task GP, making this problem a widely studied multi-task problem \cite{alvarez2011computationally, wilson2012gaussian}.

Mean absolute error (MAE) and running time are used as criterion to compare all existing algorithms. Following the previous works, we also restart the experiment 10 times with various initializations of parameters and average the MAE. Note that in our model, the initialization for "weights" are predetermined to be a set of orthonormal delta vectors, only the initialization of GP-related parameters are free to change. Experimental setup follows \cite{goovaerts1997geostatistics, alvarez2011computationally, wilson2012gaussian} and the results are displayed in Table \ref{JuraTable}. Our results are denoted by EMGPR and results from other algorithms are from \cite{wilson2012gaussian}. To compare the training time and to make sure it is not influenced by the CPUs of workstations used, we start by implementing the "no transfer" GP approach. The times reported in Table \ref{JuraTable} for training the EMGPR are normalized by the time needed for the no transfer approach. The reported training time in \cite{wilson2012gaussian} are also normalized by the time needed of their no transfer implementation. Following the discovery in \cite{wilson2012gaussian} that metal concentration data is not Gaussian distributed, each task is first log transformed before normalized to zero mean and unit variance. We also follow the notation used in \cite{wilson2012gaussian} that $*$ corresponds to results on untransformed data. 

As we notice from the results displayed in Table \ref{JuraTable}, our EMGPR beats all others with improved MAE and reduced computation time on transformed data. On untransformed data, our approach achieves comparable MAE w.r.t the best existing approach-GPRN(VB) but with greatly reduced computation time. We also notice from the results that the ensemble learning algorithm not only reduces computation time but also increases the regression performance, especially for the transformed data. Note, our approach is also extremely robust to the initialization of $\boldsymbol{\alpha}_i$s and $\sigma_i$s.

\begin{table}[!t]
	\caption{JURA Cadmium concentration prediction}
	\label{JuraTable}
	\centering
	\begin{tabular}{llllll}
		\toprule
		Approach & Average MAE & Time & Approach & Average MAE & Time \\
		\midrule
		EMGPR (Ensemble) & \textbf{0.4025 (3.66e-7)} & \textbf{2.16} & GP &0.5739 (0.0003) & 1.00 \\
		EMGPR & 0.4212 (8.31e-10) & 2.96 & ICM & 0.4608 (0.0025) & 6.85 \\
		EMGPR* (Ensemble) & 0.4531 (2.37e-5) & 2.15 & CMOGP & 0.4552 (0.0013) & 10.59 \\
		EMGPR* & 0.4602 (1.47e-9) & 3.06 & Co-kriging & 0.51 & \\
		GPRN (VB)& 0.4040 (0.0006) & 14.05 & SLFM & 0.4578 (0.0025) & 10.70\\
		GPRN* (VB) & 0.4525 (0.0036) & 16.08 & SLFM  (VB) & 0.4247 (0.0004) & 8.30 \\
		&&& SLFM* (VB) & 0.4679 (0.0030) & 10.95 \\
		\bottomrule
	\end{tabular}
\end{table}

To explore how the ensemble and the size of mini-batches influence the regression result, we performed another set of  regression experiments, in which, instead of estimating only the concentration of Cadmium at test sites with additional information from zinc and nickel, here we estimate all three chemicals at each test sites. The "no transfer" GP model is implemented as a baseline algorithm and the ICM model from a highly optimized GPy implementation \cite{gpy2014} is also used as a reference. Following the setup for ICM in \cite{alvarez2011computationally, wilson2012gaussian}, which allows for the best performance of ICM, we set $R_1 = 2$, meaning that the $\boldsymbol{B}$ used in the ICM is of rank 2.

\begin{figure}[!tb]
	\begin{minipage}{0.48\textwidth}
		\centering
		\includegraphics[width=1.1\linewidth]{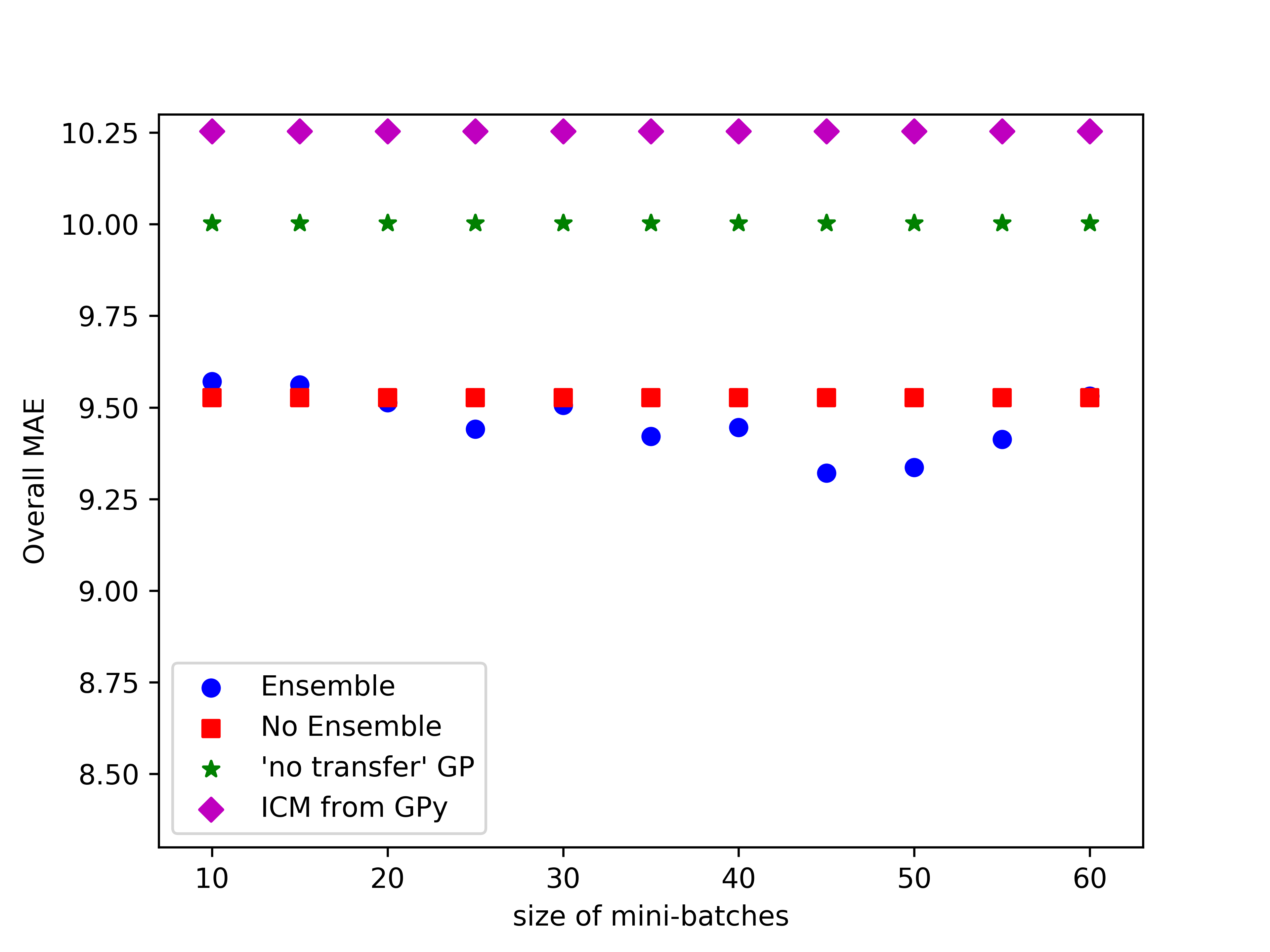}
		\caption{A comparison of overall MAE as the size of mini-batches changes}\label{plotOverallMAE}
	\end{minipage}\hfill
	\begin {minipage}{0.48\textwidth}
	\centering
	\includegraphics[width=1.1\linewidth]{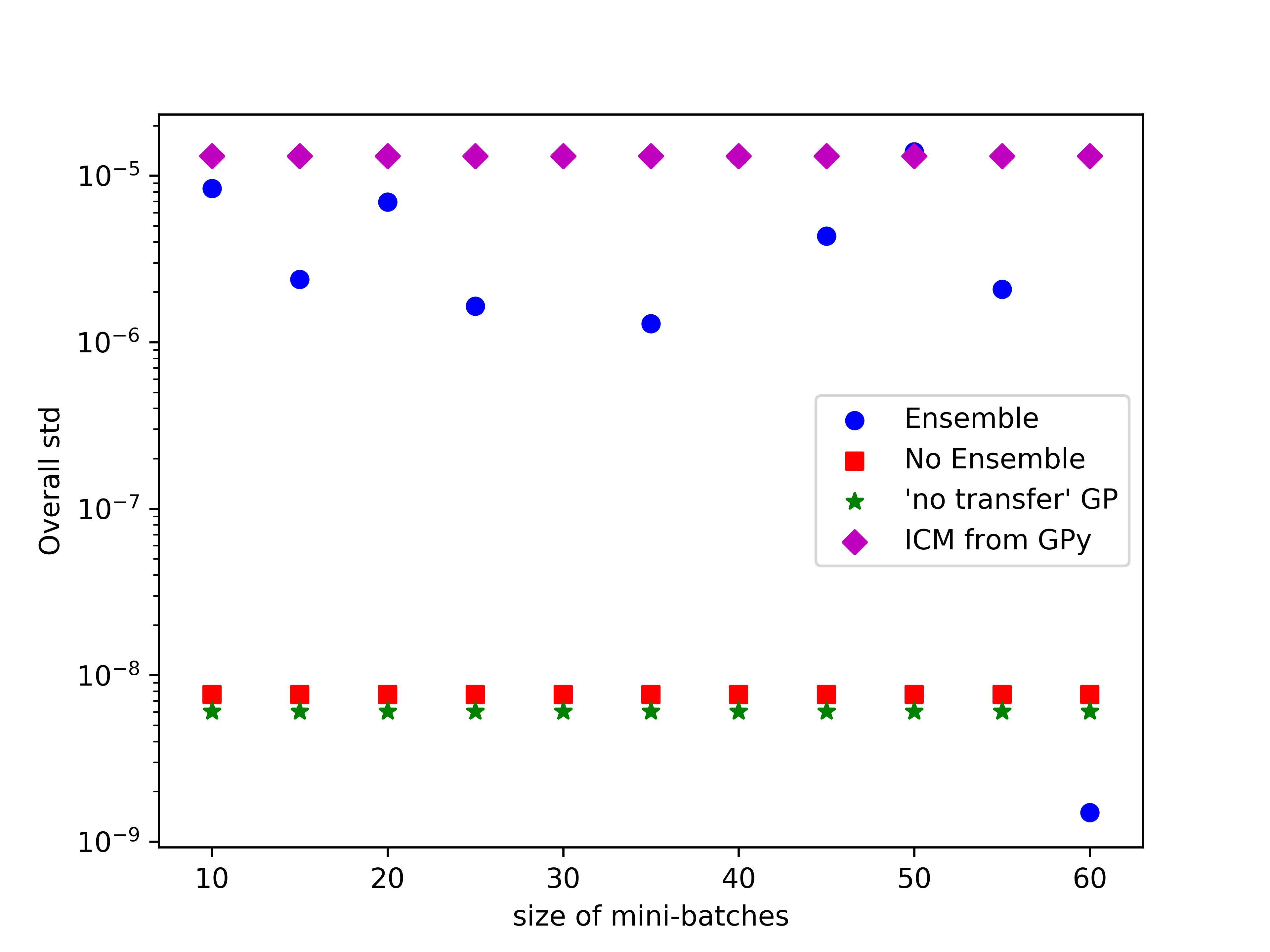}
	\caption{A comparison of overall standard deviation (in log scale) as the size of mini-batches changes }\label{plotOverallSTD}
\end{minipage}
\end{figure}

The experimental setup is similar to a standard regression but instead of a scalar value output, a vector value output is expected at each testing site. Data are first log transformed and normalized. We also use the overall MAE and standard deviation(SD) as criterion to compare all algorithms and each algorithm is repeated 10 times with different initializations. The results are displayed in both Figure \ref{plotOverallMAE} and Figure \ref{plotOverallSTD}, where the SD are plotted in log scale in Figure \ref{plotOverallSTD}. From both plots, the baseline algorithm "no transfer" GP performs even better than the ICM with reduced MAE and SD. Our model without the ensemble yields further reduced MAE with SD almost as small as the "no transfer" GP. If we add ensemble and properly choose the size of mini-batches, we can further reduce the MAE but at a cost of a relatively higher SD. However, as seen in the supplemental material, for Cadmium and Nickel, ensemble training does degrade the MAE slightly but the increase in both cases is much smaller compared with the decrease in Zinc and this is the reason the overall MAE decreases.

We notice that, by the use of ensemble, we trade the stability of our model to achieve better performance. For most of the time, ensemble would yield smaller MAE with much higher SD. However, since the SD is $10^6$ times smaller compared with the MAE, the randomness induced by  the ensemble is trivial compared to the reduction in MAE. Also, even after using the ensemble, the SD of our algorithm is smaller than that of the ICM method. Another observation from Figure \ref{plotOverallMAE} is there is an optimal value (45 in our experiment) for the  size of mini-batches such that the MAE is the smallest. This can be viewed as a trade off between the number of ensembles and the size of each mini-batches. Ideally, we prefer more ensembles to provide a more averaged result, however this would result in smaller mini-batches, which would potentially harm the parameter estimation. Additional plots of MAE and SD that corresponds to each task (chemical) can be found in the supplemental material.

\subsection{NOAA ATMS Dataset}

This dataset is acquired on June 1, 2014, and served as a part of NASA's ATMS Temperature Data Record from the NOAA CLASS database. It contains ATMS soundings from 10 out of 22 channel bands at 1080 geolocations (2D coordinates). A common task with ATMS data is to infer a vector of unobserved data at locations from a given grid. Since we have no way to test inference results on unobserved locations, in the following experiments, we simply divide the dataset into training and testing and compare the performance using the known testing dataset. Hence, the objective of this experiment is to predict vector value outputs on test sites given data from training sites. This problem can be easily modeled as a multi-task problem where data from each channel bands corresponds to a specific task and the coordinates are used as 2D inputs. In terms of multi-task GP, data from each channel bands are modeled using a GP and all GPs share the same inputs (2D coordinates).

We performed a set of 10 experiments, during each experiment, we randomly pick 1000 data points and used $80 \%$ of those (800) as training and the remaining $20 \%$ as testing. A five-fold cross validation (CV) is also used to avoid extreme cases. In these experiments, the mean squared error (MSE) and MAE are used as criterion to compare all algorithms. The baseline algorithm that we compare with is the "no transfer" GP model where each task (GP learning and regression) is implemented using the GP module from \textit{scikit-learn} \cite{scikit-learn} and the ICM model from GPy \cite{gpy2014} with $R_1 = 1, 2 \text{ and } 3$. For all algorithms, each GP or latent process uses the same initialization for the length-scale and noise variance ($\alpha_{d,0} = 1.0$ and $\sigma_{d,0}^2 = 0.02$, $\forall d = 1, 2, \cdots, D$), which are reasonable due to the fact that data from each task has been normalized.

The average MSE with SD from all 10 experiments are tabulated in Table \ref{ATMStableMSE}. In terms of the overall performance, our EMGPR yields a performance 17.9\% better compared with the "no transfer" GP, 12.7\% better with the ICM($R_1 = 1$), 21\% better with the ICM($R_1 = 2$) and 23.7\% better with the ICM($R_1=3$). In addition to the average MSE, our EMGPR also generates the lowest SD among all five tested algorithms. For individual tasks, our EMGPR also outperforms all other tested algorithms on almost every task except task \#2, where, in terms of the MSE, we perform only slighted worse compared with the "no transfer" GP, but with a reduced SD. 

The average MAE with SD are displayed in Table \ref{ATMStableMAE}. Due to limited page length, we only present the overall MAE in the paper and the task-specific MAE are placed in the supplemental material. In terms MAE, as with Table \ref{ATMStableMSE}, $R_1 = 1$ is the best ICM model, however, the "no transfer" GP performs even better and again, our EMGPR beats both approaches by 8.8\% and 7.5\% respectively. Note, although results in Table \ref{ATMStableMSE} and Table \ref{ATMStableMAE} are averaged result, we notice that our EMGPR outperforms all other tested algorithms in every experiment.

\begin{table*}[!t]
	\renewcommand{\arraystretch}{1.0}
	\small
	\tabcolsep=0.03cm
	\caption{A comparison of average MSE}
	\label{ATMStableMSE}
	\centering
	\begin{tabular}{ccccccccccc}
		\toprule
		 Task \# & GP & ICM($R_1 = 1$) & ICM($R_1 = 2$) & ICM($R_1 = 3$) & EMGPR \\
		 \midrule
		   1 & 0.01378(2.900e-3)   & 0.01997(4.508e-3)   & 0.01826(2.897e-3)   & 0.01884(3.424e-3) & \textbf{0.01165(1.901e-3)} \\
		   2 & \textbf{0.02943(4.139e-3)}   & 0.04274(5.515e-3)   & 0.03664(3.015e-3)   & 0.03428(3.879e-3) & 0.03079(2.927e-3) \\
		   3 & 0.06639(3.187e-3)   & 0.06234(7.051e-3)   & 0.06520(4.487e-3)   & 0.06731(6.486e-3) & \textbf{0.04148(1.886e-3)} \\
		   4 & 0.05242(1.830e-3)   & 0.05080(5.872e-3)   & 0.05447(3.290e-3)   & 0.05754(5.590e-3) & \textbf{0.03568(1.547e-3)} \\
		   5 & 0.03870(1.155e-3)   & 0.04146(3.970e-3)   & 0.05086(4.873e-3)   & 0.05480(5.207e-3) & \textbf{0.03216(1.048e-3)} \\
		   6 & 0.12457(4.992e-3)   & 0.11475(8.788e-3)   & 0.11353(6.507e-3)   & 0.11570(7.833e-3) & \textbf{0.10417(5.194e-3)} \\
		   7 & 0.28401(33.020e-3) & 0.23899(24.349e-3) & 0.26272(23.377e-3) & 0.26782(28.845e-3) & \textbf{0.23624(16.755e-3)} \\
		   8 & 0.09222(12.407e-3) & 0.08303(7.222e-3)   & 0.09555(9.582e-3)   & 0.10062(12.904e-3) & \textbf{0.07670(5.009e-3)} \\
		   9 & 0.03839(5.154e-3)   & 0.03949(2.200e-3)   & 0.05586(7.724e-3)   & 0.06017(8.021e-3) & \textbf{0.03550(1.664e-3)} \\
		 10 & 0.02550(2.255e-3)   & 0.02611(1.675e-3)   & 0.04250(6.242e-3)   & 0.04580(5.508e-3) & \textbf{0.02382(1.089e-3)} \\
		 \midrule
		 Overall & 0.07654(4.416e-3) & 0.07197(4.899e-3) & 0.07956(3.075e-3) & 0.08229(4.228e-3) & \textbf{0.06282(2.667e-3)}\\
		 Improve & \textbf{17.9\%} & \textbf{12.7\%} & \textbf{21.0\%} & \textbf{23.7\%} & NA\\
		\bottomrule
		
	\end{tabular}
\end{table*}

\begin{table*}[!t]
	\renewcommand{\arraystretch}{1.0}
	\small
	\tabcolsep=0.08cm
	\caption{A comparison of average MAE}
	\label{ATMStableMAE}
	\centering
	\begin{tabular}{cccccc}
		\toprule
				 & GP & ICM($R_1 = 1$) & ICM($R_1 = 2$) & ICM($R_1 = 3$) & EMGPR \\
				 \midrule
		Overall & 0.14666(1.535e-3) & 0.14870(2.356e-3) & 0.15608(2.468e-3) & 0.15938(3.368e-3) & \textbf{0.13566(1.696e-3)} \\
		Improve & \textbf{7.5\%} & \textbf{8.8\%} & \textbf{13.1\%} & \textbf{14.9\%} & NA\\

		\bottomrule
		
	\end{tabular}
\end{table*}

\section{Conclusions}

In this paper, we propose to solve the multi-task GP problem by using a new model which learns a mixture of multiple latent processes through the decomposition of the covariance matrix into a sum of Kronecker products combining "weights" and spatial covariance. From this sum structure, we propose a parallelizable parameter learning algorithm with a determined initialization. We also notice that by using an ensemble learning instead of batch learning, we trade variance for accuracy. However, even after the use of ensemble, the variance of our model is still smaller compared with the ICM model, but with a boosted accuracy. Another advantage of our parameter learning algorithm is that it can easily incorporate latest GP algorithms without the need of designing a whole new multi-task GP framework from the ideas used in latest GP algorithms.

%

\newpage

%


%

\bibliography{refs.bib}

\end{document}



\maketitle


This supplemental material is organized as follows: We first give a review of Gaussian Process Regression (GPR), Multi-task GPR with sum of separable kernels. Then we explain in details how our model is originated and optimized from the "no transfer" GP model. A detailed the gradient calculation needed in our parameter estimation algorithm is also presented. In the end, more results are included.

\section{A Quick Review of GPR and Multi-task GPR}

\subsection{Gaussian Process Regression (GPR)}

A traditional regression task requires the learning of a scalar-value function $f(x)$, given a set of $N$ $P$-dimensional training inputs $\{\boldsymbol{x}_1, \cdots, \boldsymbol{x}_N\footnote{Throughout this material, we use bold symbols for both matrices (capitalized) and vectors (lower case).}\}$, where $\boldsymbol{x}_i = \begin{bmatrix}
x_1 & \dotsb & x_P 
\end{bmatrix}^T \in \mathbb{R}^P$, and their corresponding output $\{ f(\boldsymbol{x}_1), \cdots, f(\boldsymbol{x}_N) \}$. Here we denote a vector of training outputs by $\boldsymbol{f} = \begin{bmatrix}
f(\boldsymbol{x}_1) & \dotsb & f(\boldsymbol{x}_N)
\end{bmatrix}^T \in \mathbb{R}^N$. From the perspective of Gaussian Process (GP), the function $f(x)$ is modeled with a GP prior,
\begin{equation} \label{GPprior}
f \sim \mathcal{GP} \left( \mu \left( \boldsymbol{x} \right), k \left( \boldsymbol{x}, \boldsymbol{x}' \right) \right),
\end{equation}
resulting in a joint Gaussian distribution for function outputs,
\begin{equation} \label{GPdist}
\boldsymbol{f} \sim \mathcal{N} \left( \boldsymbol{\mu}, \boldsymbol{K_f} \right),
\end{equation}
where $\boldsymbol{\mu}$ is a mean vector of size $N$, $\boldsymbol{K_f}$ is a covariance matrix of size $N \times N$ the $i,j$-th element of $\boldsymbol{K_f}$, $k_{ij} = k \left( \boldsymbol{x}_i, \boldsymbol{x}_j \right)$. Note, for simplicity, $\mu \left( \boldsymbol{x} \right)$ is always assumed to be 0, with $\boldsymbol{\mu}$ being a zero vector $\boldsymbol{0}$. This assumption can always be achieved by centering the data to zero mean.

The function $k \left( \boldsymbol{x}, \boldsymbol{x}' \right)$ in equation \eqref{GPprior} is referred to as either the covariance function or the kernel function, which in many cases, including the work in this paper, is taken as a Squared Exponential (SE) or RBF kernel,
\begin{equation} \label{SEkernel}
k \left( \boldsymbol{x}, \boldsymbol{x}' \right) = \exp \left( \frac{1}{2 \alpha^2} \lVert \boldsymbol{x} - \boldsymbol{x}' \rVert^2_2 \right).
\end{equation}
The parameter $\alpha$ sets the characteristic length scale of the correlation structure. This function can also be extended to allow separate length-scales for each input dimension, resulting in the automatic relevance determination (ARD) function of the form \cite{williams2006gaussian}:
\begin{equation} \label{ARDkernel}
k \left( \boldsymbol{x}, \boldsymbol{x}' \right) = \exp \left(  \sum_{i=1}^{P}\frac{1}{2 \alpha_i^2} \left( x_i - x_i'  \right)^2  \right).
\end{equation}
In both equation \eqref{SEkernel} and \eqref{ARDkernel}, the variance is assumed to be unity which can be easily established through standard normalization of the input data.

To account for noise, we use $\boldsymbol{y}$ to denote a noisy version of $\boldsymbol{f}$ and from equation \eqref{GPdist}, the joint distribution for $\boldsymbol{y}$ can be written as:
\begin{equation}
\boldsymbol{y} \sim \mathcal{N} \left( \boldsymbol{0}, \boldsymbol{K_f} + \sigma^2 \boldsymbol{I} \right),
\end{equation}
where $\sigma^2$ denotes the noise variance and $\boldsymbol{I}$ is an identity matrix with the same size of $\boldsymbol{K_f}$.

According to the GPR framework, when a set of test inputs $\boldsymbol{X}^*$ is given, the predictive distribution of their corresponding outputs, $\hat{\boldsymbol{f}}^*$, also follows a Gaussian distribution,
\begin{equation}
\hat{\boldsymbol{f}}^* \sim \mathcal{N} \left( \boldsymbol{K}_{\boldsymbol{f}^*, \boldsymbol{f}} \left( \boldsymbol{K_f} + \sigma^2 \boldsymbol{I} \right)^{-1} \boldsymbol{y}, \boldsymbol{K}_{\boldsymbol{f}^*} - \boldsymbol{K}_{\boldsymbol{f}^*, \boldsymbol{f}} \left( \boldsymbol{K_f} + \sigma^2 \boldsymbol{I} \right)^{-1} \boldsymbol{K}_{\boldsymbol{f}, \boldsymbol{f}^*} \right),
\end{equation}
where $\boldsymbol{K}_{\boldsymbol{f}^*}$ is the autocovariance matrix of $\boldsymbol{f}^*$ and $\boldsymbol{K}_{\boldsymbol{f}^*, \boldsymbol{f}}$, $\boldsymbol{K}_{\boldsymbol{f}, \boldsymbol{f}^*}$ are crosscovariance matrices between $\boldsymbol{f}^*$ and $\boldsymbol{f}$.

\subsection{Multi-task GPR with Sum of Separable (SoS) Kernels}

In multi-task GPR, the function that needs to be learned expands from a 1-dimensional space into a $D$-dimensional space, where for a given inputs, $\boldsymbol{x}_i$, the output becomes a vector $\begin{bmatrix}
f_1(\boldsymbol{x}_i) & \dotsb & f_D(\boldsymbol{x}_i
\end{bmatrix}^T$ of size $D$. Here $f_d(\boldsymbol{x}_i)$ stands for the $d$-th output or task of input $\boldsymbol{x}_i$. In Geostatistics, a "linear model of coregionalization" (LMC) model is often applied to this type of data. In machine learning community, Alvarez et al refers to the type of kernel functions used in LMC as SoS kernels \cite{alvarez2012kernels}.

In LMC, the $d$-th output of any input $\boldsymbol{x}$ is expressed as a linear combination of $Q$ latent functions:
\begin{equation} \label{LMCBasic}
f_d(\boldsymbol{x}) = \sum_{q=1}^{Q} a_{d,q} u_q(\boldsymbol{x}),
\end{equation}
where each $u_q(\boldsymbol{x})$ stands for a latent function with different and independent GP priors. Here each latent function arises from a unique and independent latent GP. This linear mixture can also be extended by allowing multiple independently distributed latent functions from the same latent process,
\begin{equation} \label{LMCExtended}
f_d(\boldsymbol{x}) = \sum_{q=1}^{Q} \sum_{l=1}^{R_q} a_{d,q}^l u^l_q(\boldsymbol{x}),
\end{equation}
where $\{ u^l_q(\boldsymbol{x}) \} _{l=1}^{R_d}$ share the same GP prior and $R_d$ denotes the total number of latent functions that have the same GP prior. Note that even $\{ u^l_q(\boldsymbol{x}) \} _{l=1}^{R_d}$ have the same GP prior, they are still independent. From equation \eqref{LMCExtended}, with the independence assumption, the covariance between $f_d(\boldsymbol{x})$ and $f_{d'}(\boldsymbol{x}')$ can be expressed as:
\begin{equation} \label{LMCCovariance}
\text{cov} \left( f_d(\boldsymbol{x}), f_{d'}(\boldsymbol{x}') \right) = \sum_{q=1}^{Q} \sum_{l=1}^{R_q}  a_{d,q}^l a_{d',q}^{l} \text{cov} \left( u^l_q(\boldsymbol{x}), u^l_q(\boldsymbol{x}) \right) = \sum_{q=1}^{Q} b_{d, d'}^q \text{cov} \left( u^l_q(\boldsymbol{x}), u^l_q(\boldsymbol{x}) \right).
\end{equation}
The kernel can also be easily derived as
\begin{equation} \label{LMCSoS}
\boldsymbol{K} \left( \boldsymbol{x}, \boldsymbol{x}' \right) = \sum_{q=1}^{Q} \boldsymbol{B}_q k_q \left( \boldsymbol{x}, \boldsymbol{x}' \right)
\end{equation}
where the $\boldsymbol{B}_q$ is a coregionalization matrix of size $D \times D$ and the $d, d'$-th element of $\boldsymbol{B}_q$ is $b_{d,d'}^q$. Note we abuse the notation here a little by using a capitalized bold letter $\boldsymbol{K}$ to denote a matrix-value function instead of a constant matrix. From equation \eqref{LMCSoS}, this SoS kernel is a mixture of $Q$ different kernels. From the perspective of latent processes, this SoS kernel corresponds to an assumption that outputs are from a mixture of $Q$ latent processes.

A special case of the LMC model is when $Q = 1$, resulting in a simplified kernel of the form,
\begin{equation} \label{ICMkernel}
\boldsymbol{K} \left( \boldsymbol{x}, \boldsymbol{x}' \right) =  \boldsymbol{B} \cdot k \left( \boldsymbol{x}, \boldsymbol{x}' \right),
\end{equation}
which is widely used in the machine learning community.

\section{Our Model and its Gradients}

\subsection{Our Model improves from "no transfer" GP}

Let the matrix $\boldsymbol{Y} = \begin{bmatrix}
\boldsymbol{y}_1 & \boldsymbol{y}_2 & \dotsb & \boldsymbol{y}_D
\end{bmatrix} \in \mathbb{R}^{N \times D}$ be a set of responses with $N$ samples and $D$ tasks, where each column $\boldsymbol{y}_d$ represents the $d$-th task with length $N$. We also use $\boldsymbol{y} = vec \left( \boldsymbol{Y} \right) = \begin{bmatrix}
\boldsymbol{y}_1^T & \dotsb & \boldsymbol{y}_D^T
\end{bmatrix}^T \in \mathbb{R}^{ND \times 1}$  and denote by $\boldsymbol{X}$ a set of $N$ inputs $\boldsymbol{x}_1, \boldsymbol{x}_2, \cdots, \boldsymbol{x}_N$. 

In the "no transfer" model, each task is modeled by an independent GP. Assuming each task is normalized with zero mean and unit variance, $\boldsymbol{y}_i$ is distributed as:
\begin{equation}
\boldsymbol{y}_i \sim \mathcal{N} \left( \boldsymbol{0}, \boldsymbol{K}_i + \sigma_i^2 \boldsymbol{I} \right) \qquad \qquad \forall i = 1, 2, \cdots, D.
\end{equation}
The resulting distribution of $\boldsymbol{y}$ is of the form:
\begin{equation} \label{Covy}
\boldsymbol{y} \sim \mathcal{N} \left( \begin{bmatrix}
\boldsymbol{0} \\
\vdots \\
\boldsymbol{0}
\end{bmatrix}, \begin{bmatrix}
\boldsymbol{K}_1 & & \boldsymbol{0} \\
 & \ddots & \\
\boldsymbol{0} & & \boldsymbol{K}_D
\end{bmatrix} + \begin{bmatrix}
\sigma_1^2\boldsymbol{I} & & \boldsymbol{0} \\
& \ddots & \\
\boldsymbol{0} & & \sigma_D^2\boldsymbol{I}
\end{bmatrix} \right).
\end{equation}
Denoting the covariance matrix of $\boldsymbol{y}$ as $\boldsymbol{K_y}$, then from the distribution described in \eqref{Covy}, we have
\begin{eqnarray}
\nonumber
\boldsymbol{K_y} &=& \begin{bmatrix}
\boldsymbol{K}_1 & & \boldsymbol{0} \\
 & \ddots & \\
\boldsymbol{0} & & \boldsymbol{0}
\end{bmatrix} + \cdots + \begin{bmatrix}
\boldsymbol{0} & & \boldsymbol{0} \\
 & \boldsymbol{K}_d & \\
\boldsymbol{0} & & \boldsymbol{0}
\end{bmatrix} + \cdots + \begin{bmatrix}
\boldsymbol{0} & & \boldsymbol{0}\\
& \ddots & \\
\boldsymbol{0} & & \boldsymbol{K}_D
\end{bmatrix} + \begin{bmatrix}
\sigma_1^2\boldsymbol{I} & & \boldsymbol{0} \\
& \ddots & \\
\boldsymbol{0} & & \sigma_D^2\boldsymbol{I}
\end{bmatrix} \\
\label{SumFormMatrix}
&=& \sum_{d=1}^{D} \boldsymbol{B}_d \otimes \boldsymbol{K}_d + \boldsymbol{D} \otimes \boldsymbol{I},
\end{eqnarray}
where $\boldsymbol{B}_d$ is a zero matrix except for the $d$-th element on the diagonal which is 1 and $\boldsymbol{D} = diag \left( \sigma_1^2, \sigma_2^2, \cdots, \sigma_D^2 \right)$  is a diagonal matrix of noise covariance for each task. With the assumption that each $\boldsymbol{B}_d$ is positive semi-definite (PSD), we can rewrite equation \eqref{SumFormMatrix} as
\begin{equation}
\boldsymbol{K_y} = \sum_{d=1}^{D} \boldsymbol{W}_d \boldsymbol{W}_d^T \otimes \boldsymbol{K}_d + \boldsymbol{D} \otimes \boldsymbol{I}
\end{equation}
with a rank-k approximatin:
\begin{equation} \label{SumFormVector}
\boldsymbol{K_y} \approx \sum_{d=1}^{D} \left( \sum_{j=1}^{k} \boldsymbol{w}_d^j \left( \boldsymbol{w}_d^j \right)^T \right) \otimes \boldsymbol{K}_d + \boldsymbol{D} \otimes \boldsymbol{I}.
\end{equation}
When only a rank-1 approximation is used, the previous approximation \eqref{SumFormVector} can be greatly simplified as
\begin{equation} \label{SumFormRank1Vector}
\boldsymbol{K_y} \approx \sum_{d=1}^{D} \left(  \boldsymbol{w}_d  \boldsymbol{w}_d^T \right) \otimes \boldsymbol{K}_d + \boldsymbol{D} \otimes \boldsymbol{I}
\end{equation}
and in the "no transfer" GP model, $\boldsymbol{w}_d = \boldsymbol{\delta}_d$, a Kronecker delta vector with the $d$-th element being 1 and zero elsewhere.

We emphasize that for the "no transfer" modek, the "weights" $\{ \boldsymbol{w}_d \}_{d=1}^D$ in \eqref{SumFormRank1Vector} are delta vectors due to the assumption that each task is generated from an independent latent process and modeled by an independent GP. This assumption prohibits the transfer of knowledge from one task to another. In order to share information across tasks, the model that we propose in this work allows $\{ \boldsymbol{w}_d \}_{d=1}^D$ to be dense vectors and they are optimized from the training dataset. Then, comparing with the "no transfer" GP model, in our model each task can be thought of as being generated from a mixture of multiple latent processes and the "weights" are controlled by these vectors $\{\boldsymbol{w}_d\}_{d=1}^{D}$.

\subsection{Gradient Calculation}

In our model, latent process related parameters $ \{\boldsymbol{\alpha}_d, \sigma_d \}_{d=1}^D$ and the "weights" $\{ \{\boldsymbol{w}_d^j\}_{j=1}^k \}_{d=1}^D$ needs to be learned from the training dataset. Since in this work, only rank-1 approximation and the SE kernel function for each latent process are used, the set of parameters can be reduced to $\{\alpha_d, \sigma_d \}_{d=1}^D$ and $\{\boldsymbol{w}_d\}_{d=1}^D$.

In our two-step parameter learning algorithm, the latent process related parameters are first learned from each task via a maximum likelihood (ML) estimation:
\begin{equation}
\hat{\alpha}_d, \hat{\sigma}_d = \arg \max_{\alpha_d, \sigma_d} -\frac{1}{2} \boldsymbol{y}_d^T \boldsymbol{K}_d^{-1} \boldsymbol{y}_d - \frac{1}{2} \log \lvert \boldsymbol{K}_d \rvert - \frac{N}{2} \log \left( 2 \pi \right)
\end{equation}
with gradient
\begin{eqnarray}
\frac{\partial}{\partial \theta} \log p \left( \boldsymbol{y}_d \lvert \boldsymbol{X}, \alpha_d, \sigma_d \right) &=& \frac{\partial}{\partial \theta}  \left( -\frac{1}{2} \boldsymbol{y}_d^T \boldsymbol{K}_d^{-1} \boldsymbol{y}_d - \frac{1}{2} \log \lvert \boldsymbol{K}_d \rvert - \frac{N}{2} \log \left( 2 \pi \right) \right) \\
&=& \frac{1}{2} \boldsymbol{y}_d^T \boldsymbol{K}_d^{-1} \frac{\partial \boldsymbol{K}_d}{\partial \theta} \boldsymbol{K}_d^{-1} \boldsymbol{y}_d - \frac{1}{2} \text{tr} \left( \boldsymbol{K}_d^{-1} \frac{\partial \boldsymbol{K}_d}{\partial \theta} \right) \\
&=& \frac{1}{2} \text{tr} \left( \left( \boldsymbol{\gamma} \boldsymbol{\gamma}^T - \boldsymbol{K}_d^{-1} \right) \frac{\partial \boldsymbol{K}_d}{\partial \theta} \right) \quad \text{where } \boldsymbol{\gamma} = \boldsymbol{K}_d^{-1} \boldsymbol{y}_d
\end{eqnarray}
and $\theta$ can be either $\alpha_d$ or $\sigma_d$. Note this gradient calculation is the same as that of the "no transfer" GP model and the standard GP.

In terms of the "weights", a similar ML estimation is utilized but maximizing the log marginal likelihood of the entire training set, $\log p \left( \boldsymbol{y} \lvert \boldsymbol{X}, \boldsymbol{w}_d, \hat{\alpha}_d, \hat{\sigma}_d \}_{d=1}^D\right)$, with $\{\hat{\alpha}_d, \hat{\sigma}_d \}_{d=1}^D$ learned from the previous step,
\begin{equation}
\{ \hat{\boldsymbol{w}}_d \}_{d=1}^D = \arg \max_{\{ \boldsymbol{w}_d \}_{d=1}^D} -\frac{1}{2} \boldsymbol{y}^T \boldsymbol{K_y}^{-1} \boldsymbol{y} - \frac{1}{2} \log \lvert \boldsymbol{K_y} \rvert - \frac{ND}{2} \log \left( 2 \pi \right).
\end{equation}
The gradient can be calculated as:
\begin{eqnarray}
\frac{\partial}{\partial \theta} \log p \left( \boldsymbol{y} \lvert \boldsymbol{X}, \{ \boldsymbol{w}_d, \hat{\boldsymbol{\alpha}}_d, \hat{\sigma}_d \}_{d=1}^D\right) &=& \frac{\partial}{\partial \theta}  \left( -\frac{1}{2} \boldsymbol{y}^T \boldsymbol{K_y}^{-1} \boldsymbol{y} - \frac{1}{2} \log \lvert \boldsymbol{K_y} \rvert - \frac{ND}{2} \log \left( 2 \pi \right) \right) \\
&=& \frac{1}{2} \boldsymbol{y}^T \boldsymbol{K_y}^{-1} \frac{\partial \boldsymbol{K_y}}{\partial \theta} \boldsymbol{K_y}^{-1} \boldsymbol{y} - \frac{1}{2} \text{tr} \left( \boldsymbol{K_y}^{-1} \frac{\partial \boldsymbol{K_y}}{\partial \theta} \right) \\
&=& \frac{1}{2} \text{tr} \left( \left( \boldsymbol{\gamma} \boldsymbol{\gamma}^T - \boldsymbol{K_y}^{-1} \right) \frac{\partial \boldsymbol{K_y}}{\partial \theta} \right) 
\end{eqnarray}
where $\boldsymbol{\gamma} = \boldsymbol{K_y}^{-1} \boldsymbol{y}$ and when $\theta$ is the $l$-th element of $\boldsymbol{w}_d$
\begin{eqnarray}
\frac{\partial \boldsymbol{K_y}}{\partial \theta}  &=& \frac{\partial}{\partial \theta} \left( \sum_{d=1}^D \boldsymbol{w}_d  \boldsymbol{w}_d^T \otimes \boldsymbol{K}_d + \boldsymbol{D} \otimes \boldsymbol{I} \right) \\
&=& \frac{\partial}{\partial \theta} \left( \boldsymbol{w}_d  \boldsymbol{w}_d^T \right) \otimes \boldsymbol{K}_d \\
&=& \left( \boldsymbol{w}_d \boldsymbol{\delta}_l^T + \boldsymbol{\delta}_l \boldsymbol{w}_d^T\right) \otimes \boldsymbol{K}_d
\end{eqnarray}

\section{More Results}

\subsection{Jura Dataset}

\begin{figure}[!tb]
	\begin{minipage}{0.48\textwidth}
		\centering
		\includegraphics[width=1.1\linewidth]{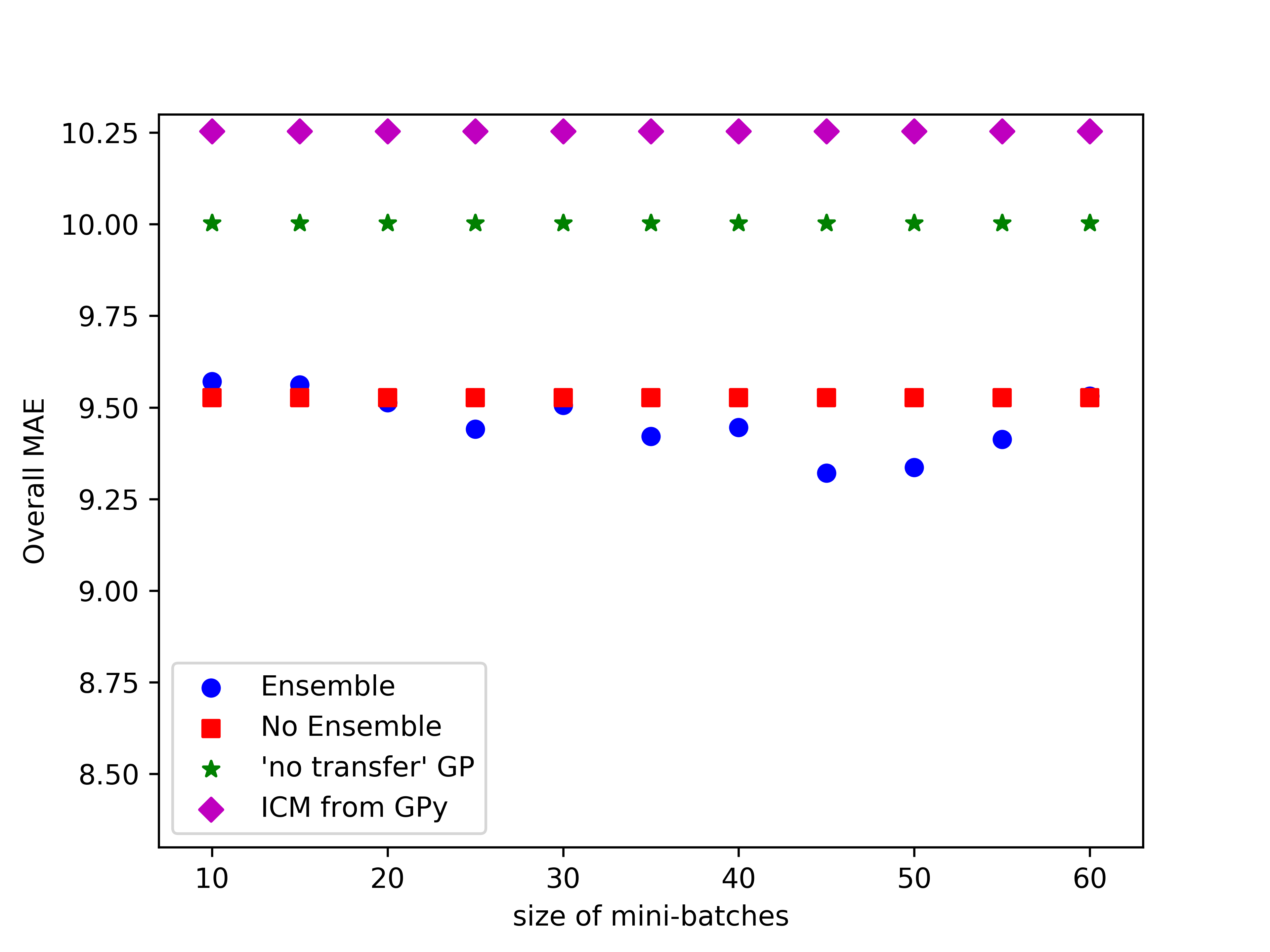}
		\caption{A comparison of overall MAE as the size of mini-batches changes}\label{plotOverallMAE}
	\end{minipage}\hfill
	\begin {minipage}{0.48\textwidth}
	\centering
	\includegraphics[width=1.1\linewidth]{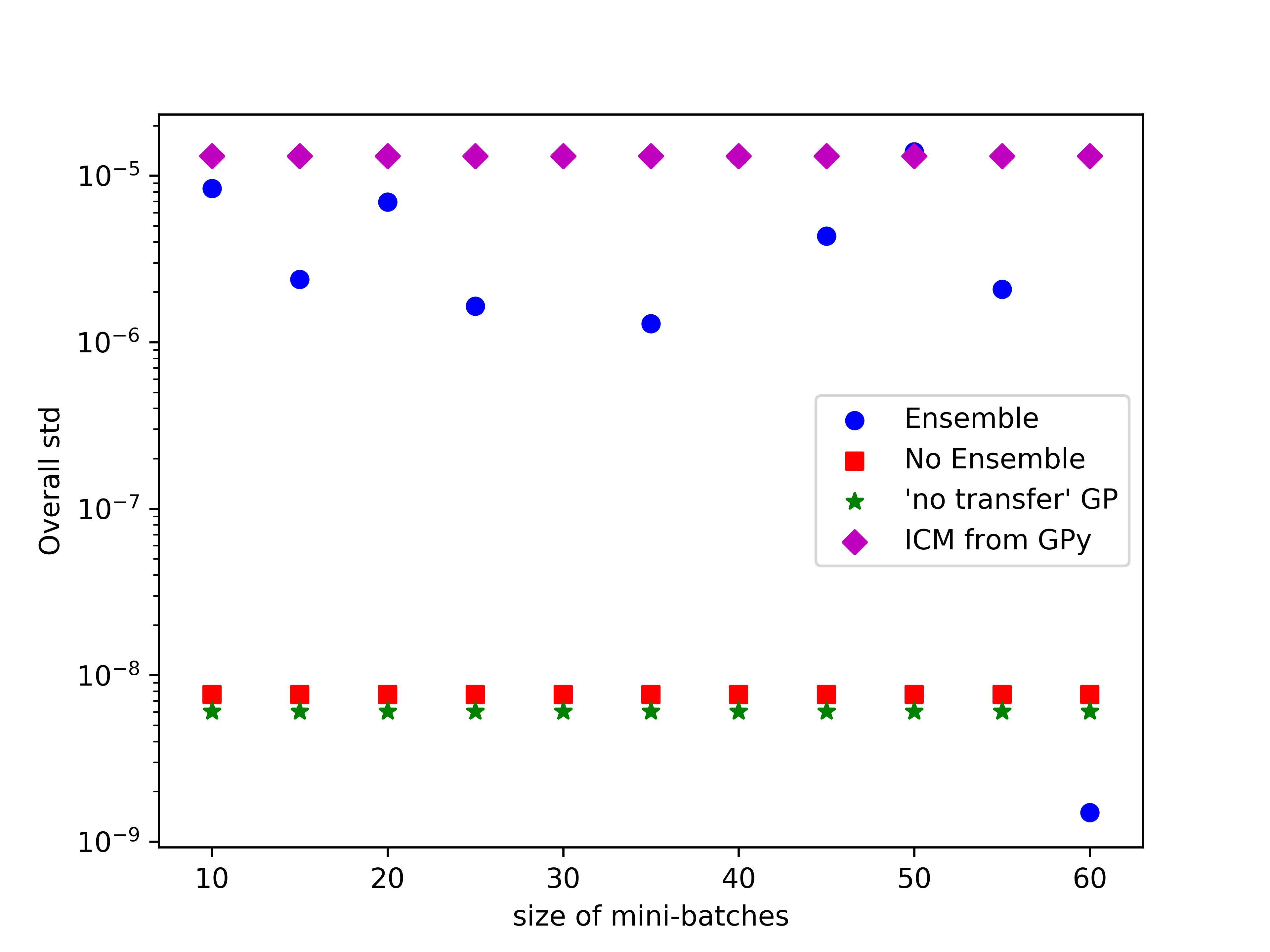}
	\caption{A comparison of overall standard deviation (in log scale) as the size of mini-batches changes }\label{plotOverallSTD}
\end{minipage}
\end{figure}

\begin{figure}[!tb]
\begin{minipage}{0.48\textwidth}
	\centering
	\includegraphics[width=1.1\linewidth]{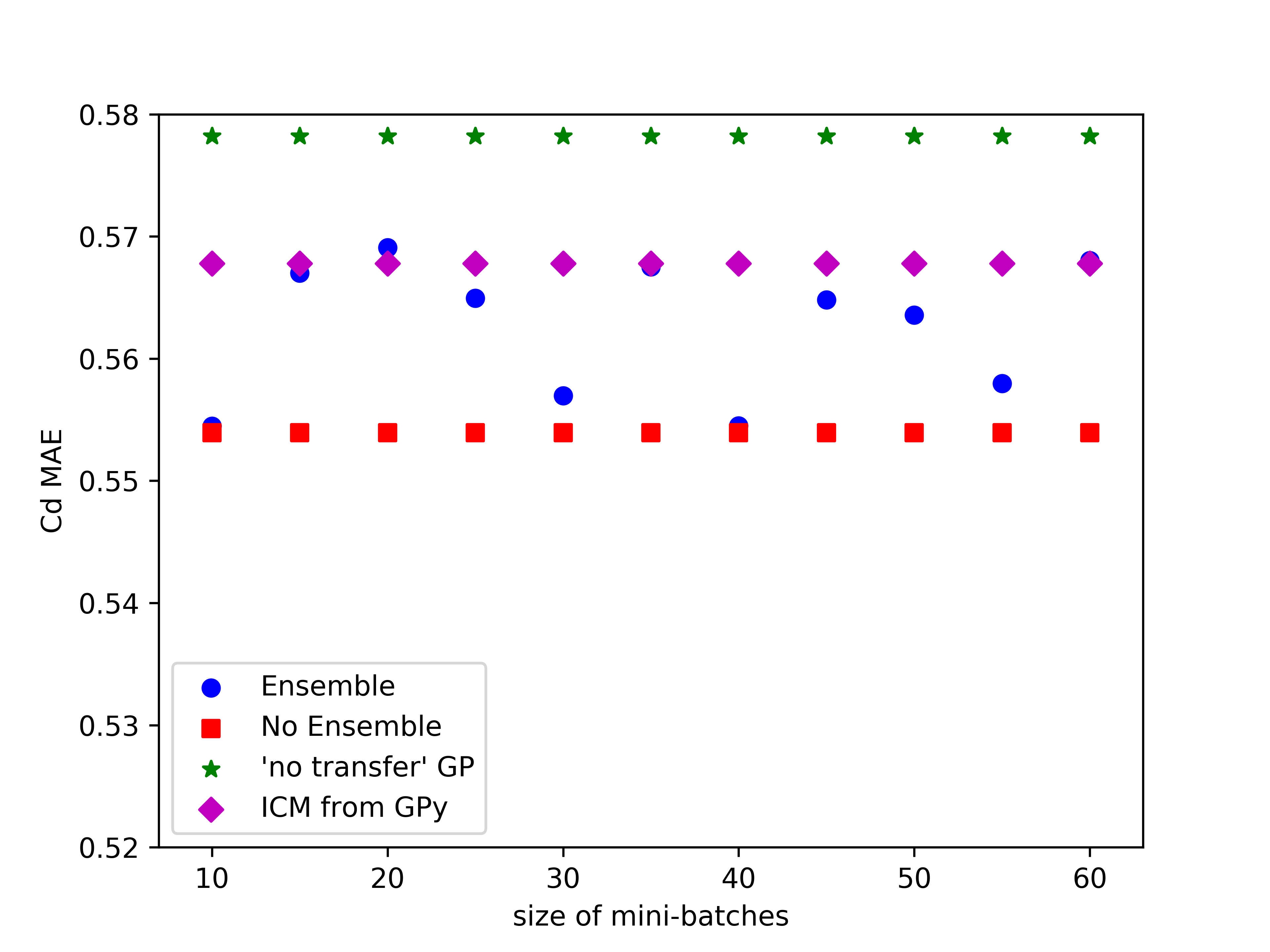}
	\caption{A comparison of Cd MAE as the size of mini-batches changes}\label{plotCdMAE}
\end{minipage}\hfill
\begin {minipage}{0.48\textwidth}
\centering
\includegraphics[width=1.1\linewidth]{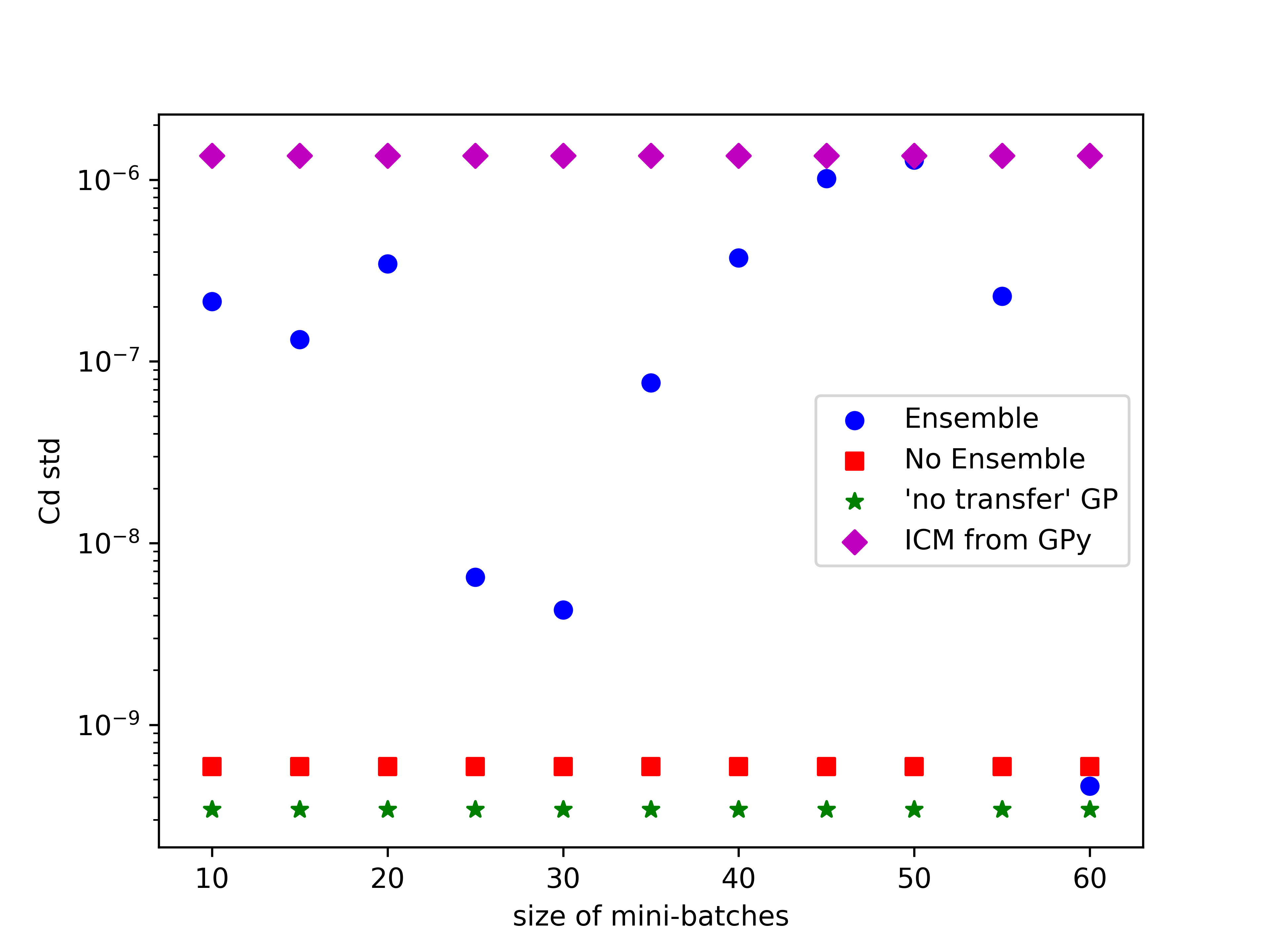}
\caption{A comparison of Cd standard deviation (in log scale) as the size of mini-batches changes }\label{plotCdSTD}
\end{minipage}
\end{figure}

\begin{figure}[!tb]
\begin{minipage}{0.48\textwidth}
\centering
\includegraphics[width=1.1\linewidth]{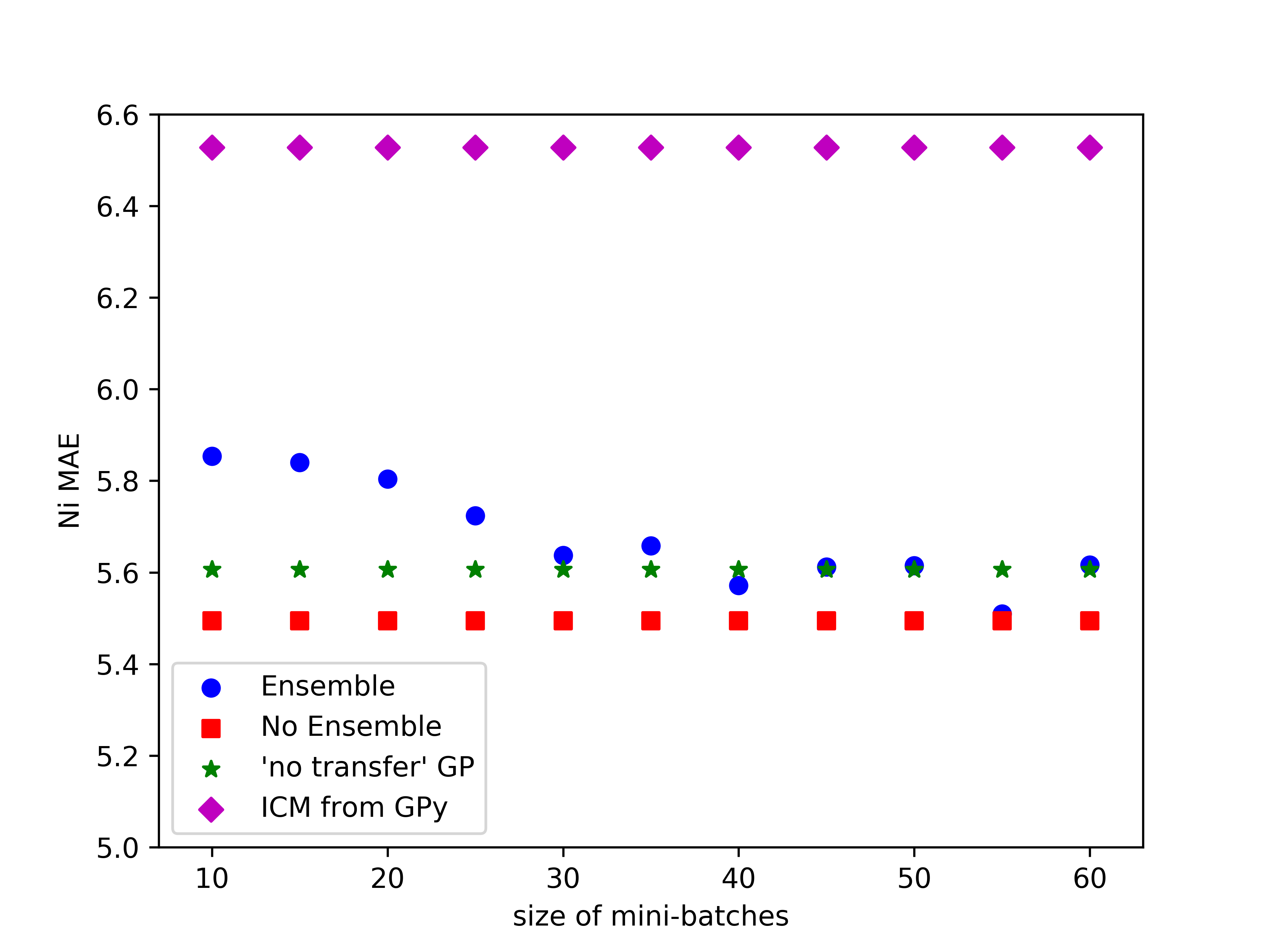}
\caption{A comparison of Ni MAE as the size of mini-batches changes}\label{plotNiMAE}
\end{minipage}\hfill
\begin {minipage}{0.48\textwidth}
\centering
\includegraphics[width=1.1\linewidth]{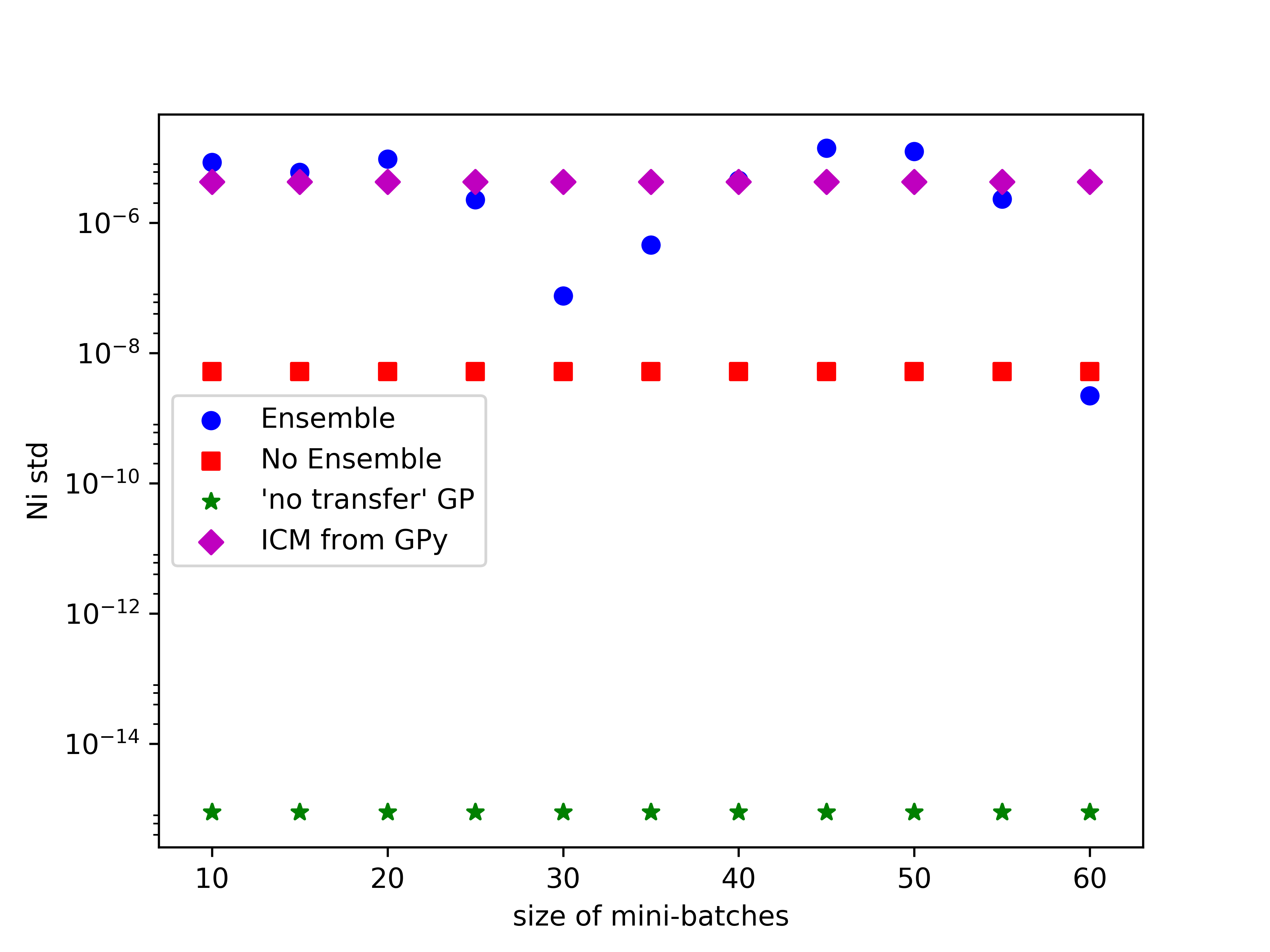}
\caption{A comparison of Ni standard deviation (in log scale) as the size of mini-batches changes }\label{plotNiSTD}
\end{minipage}
\end{figure}

\begin{figure}[!tb]
\begin{minipage}{0.48\textwidth}
\centering
\includegraphics[width=1.1\linewidth]{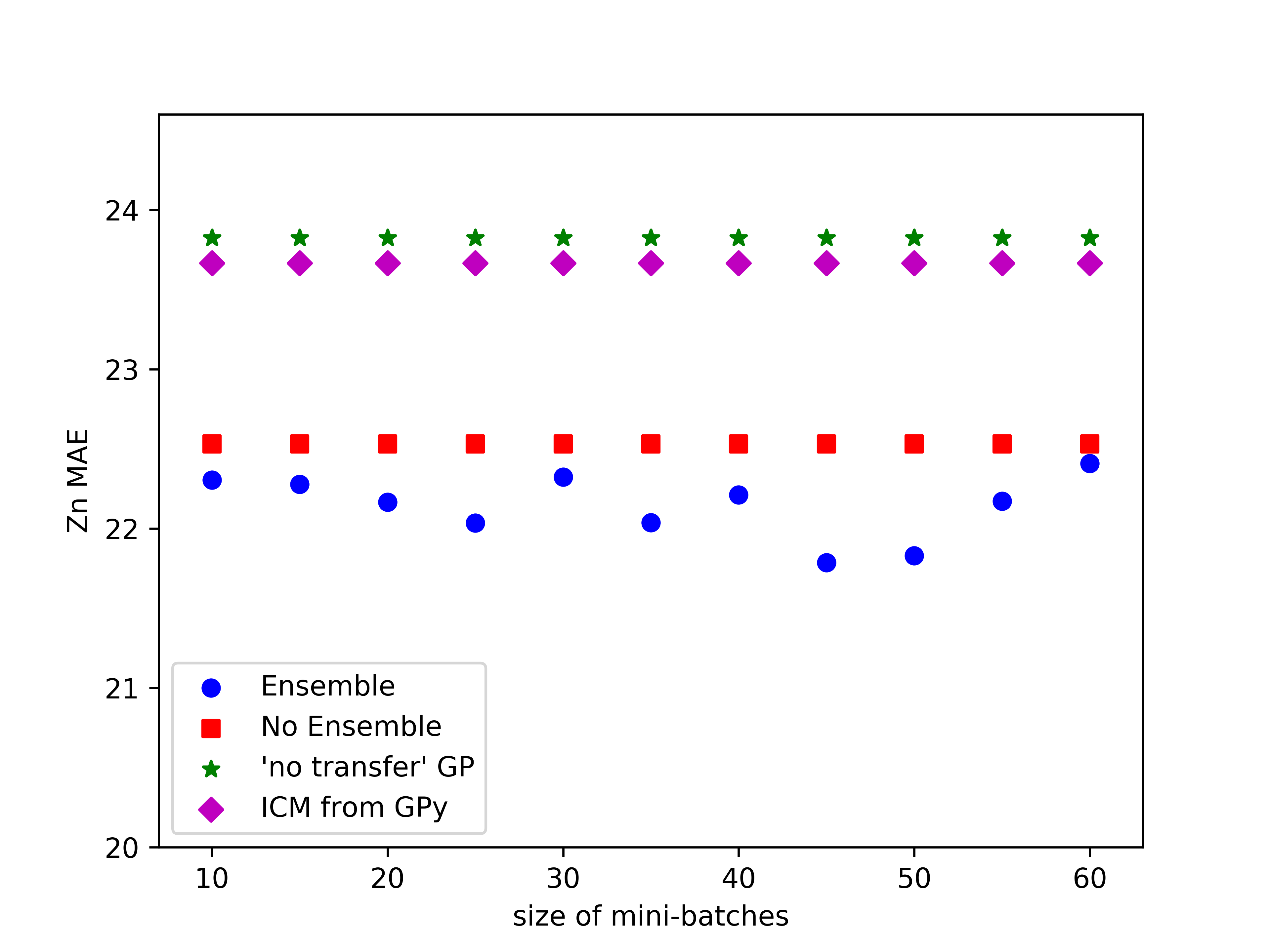}
\caption{A comparison of Zn MAE as the size of mini-batches changes}\label{plotZnMAE}
\end{minipage}\hfill
\begin {minipage}{0.48\textwidth}
\centering
\includegraphics[width=1.1\linewidth]{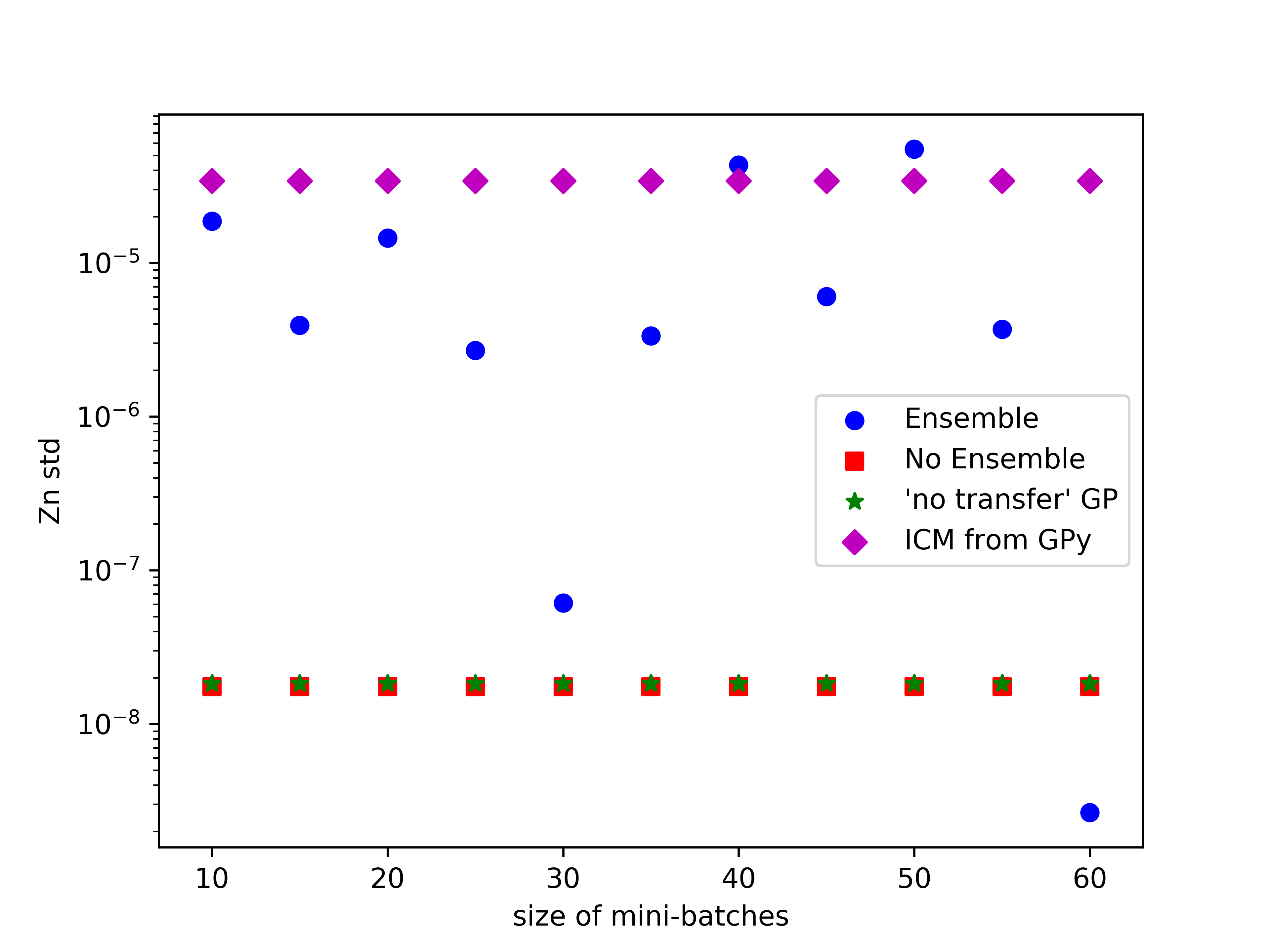}
\caption{A comparison of Zn standard deviation (in log scale) as the size of mini-batches changes }\label{plotZnSTD}
\end{minipage}
\end{figure}

Here we provide several plots of MAE and standard deviation in Figure \ref{plotOverallMAE} - \ref{plotZnSTD}, which corresponds to the overall and each task(chemical), as a function of the size of the ensemble. As mentioned in the paper, the ensemble boost the overall performance. However, we also notice that for Cadmium and Nickel, the use of the ensemble degrades the MAE, but the increase is much smaller compared with the decrease in Zinc and this is the reason the overall MAE decreases.

\subsection{NOAA ATMS Dataset}

We provide a complete table that contains task specific average MAE and its corresponding standard deviation in Table \ref{ATMStableMAE}. It's obvious from the results that, for all tasks, except one, our EMGPR outperforms the other four.

%

%


\newpage
\bibliography{refs_supp.bib}

\newpage

\newpage
\begin{table*}[!t]
	\renewcommand{\arraystretch}{1.3}
	\small
	\tabcolsep=0.08cm
	\caption{A comparison of average MAE}
	\label{ATMStableMAE}
	\centering
	\begin{tabular}{cccccc}
		\toprule
		Task \# & GP & ICM($R_1 = 1$) & ICM($R_1 = 2$) & ICM($R_1 = 3$) & EMGPR \\
		\midrule
		1 & 0.07027(3.188e-3) & 0.08172(2.777e-3) & 0.07937(3.556e-3) & 0.07971(3.851e-3) & \textbf{0.06788(2.877e-3)}\\
		2 & 0.09937(2.825e-3) & 0.11809(3.022e-3) & 0.11225(1.917e-3) & 0.10923(2.919e-3) & \textbf{0.09724(2.981e-3)}\\
		3 & 0.16000(2.337e-3) & 0.15742(4.240e-3) & 0.16589(3.879e-3) & 0.16951(6.050e-3) & \textbf{0.13156(2.319e-3)}\\
		4 & 0.14549(1.933e-3) & 0.14403(4.183e-3) & 0.15434(3.172e-3) & 0.15932(5.548e-3) & \textbf{0.12521(2.578e-3)}\\
		5 & 0.13940(1.650e-3) & 0.14417(4.487e-3) & 0.15875(4.636e-3) & 0.16390(4.901e-3) & \textbf{0.13011(2.278e-3)}\\
		6 & 0.25559(5.070e-3) & 0.23874(5.599e-3) & 0.23980(4.421e-3) & 0.24266(5.983e-3) & \textbf{0.22762(3.948e-3)}\\
		7 & \textbf{0.21073(8.236e-3)} & 0.21470(4.442e-3) & 0.22053(5.338e-3) & 0.22521(10.084e-3) & 0.21100(5.076e-3)\\
		8 & 0.15102(5.596e-3) & 0.15035(3.421e-3) & 0.16001(5.534e-3) & 0.16406(7.292e-3) & \textbf{0.14104(2.843e-3)}\\
		9 & 0.12088(3.863e-3) & 0.12352(2.232e-3) & 0.13773(6.294e-3) & 0.14301(5.993e-3) & \textbf{0.11538(2.306e-3)}\\
		10 & 0.11390(2.891e-3) & 0.11422(2.359e-3) & 0.13210(5.455e-3) & 0.13716(4.600e-3) & \textbf{0.10957(2.303e-3)}\\
		\midrule
		Overall & 0.14666(1.535e-3) & 0.14870(2.356e-3) & 0.15608(2.468e-3) & 0.15938(3.368e-3) & \textbf{0.13566(1.696e-3)} \\
		Improve & \textbf{7.5\%} & \textbf{8.8\%} & \textbf{13.1\%} & \textbf{14.9\%} & NA\\

		\bottomrule
		
	\end{tabular}
\end{table*}